\documentclass[10pt,twocolumn,letterpaper]{article}

\usepackage{wacv}
\usepackage{times}
\usepackage{epsfig}
\usepackage{graphicx}
\usepackage{amsmath}
\usepackage{amssymb}
\usepackage{booktabs}

\usepackage{subcaption, booktabs, multirow, xspace, adjustbox, xcolor, soul, tabularx, placeins, appendix}

\usepackage[accsupp]{axessibility}  

\usepackage{amssymb}
\usepackage{pifont}
\newcommand{\cmark}{\ding{51}}%
\newcommand{\xmark}{\ding{55}}%

\def\wacvPaperID{1484} 

\wacvapplicationstrack 

\wacvfinalcopy 

\ifwacvfinal
\usepackage[breaklinks=true,bookmarks=false]{hyperref}
\else
\usepackage[pagebackref=true,breaklinks=true,colorlinks=true,urlcolor=magenta,linkcolor=blue,bookmarks=false]{hyperref}
\fi

\begin{document}

\title{Benchmarking Visual Localization for Autonomous Navigation}

\author{Lauri Suomela \quad Jussi Kalliola \quad Atakan Dag \quad Harry Edelman \quad Joni-Kristian Kämäräinen\\
Tampere University, Finland\\
{\tt\small \{lauri.a.suomela,\,jussi.kalliola,\,atakan.dag,\,harry.edelman,\,joni.kamarainen\}@tuni.fi}\\
}

\maketitle
\thispagestyle{empty}

\begin{abstract}
This work introduces a simulator-based benchmark for visual localization in the autonomous navigation context. The dynamic benchmark enables investigation of how variables such as the time of day, weather, and camera perspective affect the navigation performance of autonomous agents that utilize visual localization for closed-loop control.
The experimental part of the paper studies the effects of four such variables by evaluating state-of-the-art visual localization methods as part of the motion planning module of an autonomous navigation stack. The results show major variation in the suitability of the different methods for vision-based navigation.
To the authors' best knowledge, the proposed benchmark is the first to study modern visual localization methods as part of a complete navigation stack. We make the benchmark available at \href{https://github.com/lasuomela/carla_vloc_benchmark}{\textcolor{magenta}{\url{https://github.com/lasuomela/carla_vloc_benchmark}}}.

\end{abstract}


\section{Introduction}
\label{sec:intro}



\begin{figure}[t]
  \centering
   \includegraphics[width=1.0\linewidth, trim={0 0cm 0cm 4cm}]{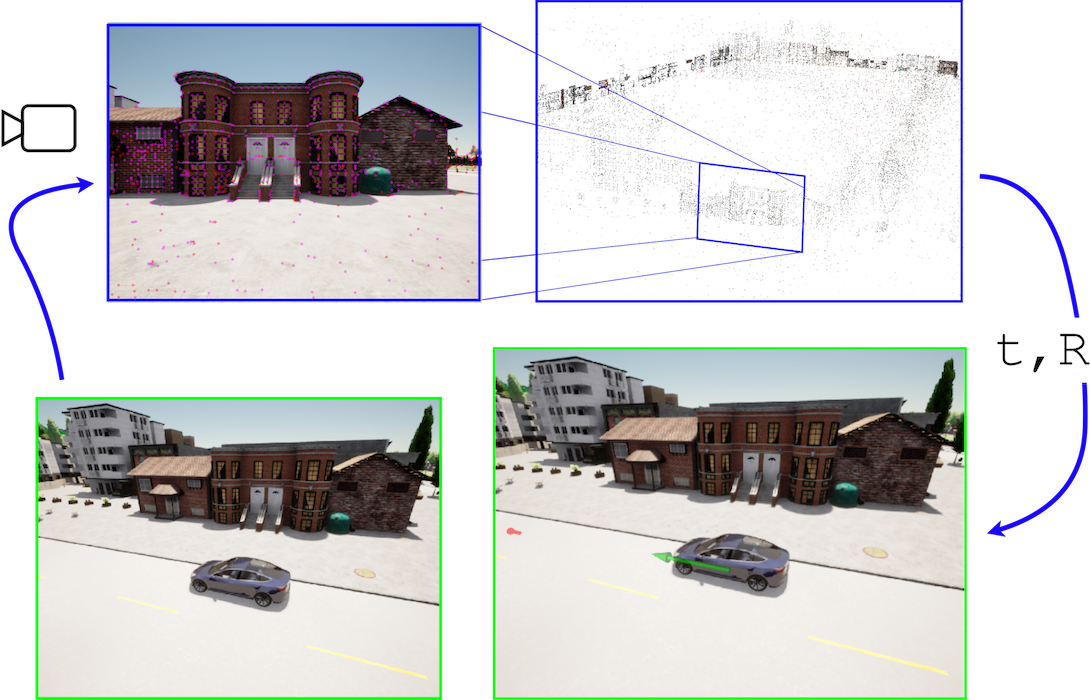}



  \caption{The vehicle finds its pose in a simulated environment by matching local features to a pre-built 3D model.} 
   \label{fig:vloc_idea}
   \vspace{-0.5cm}
\end{figure}

One of the most impressive capabilities of the human brain is the ability to take a look around, answer the question "Where am I?", and use a mental map of the environment to guide one to a place that has been visited before.
A task that seems trivial to humans is notoriously difficult for robots.
One promising approach to {\em autonomous navigation} is vision-based navigation that uses {\em visual localization}~\cite{sattler_benchmarking_2018} to estimate the pose of an agent with respect to a metric map. The map is created prior to navigation by creating a 3D reconstruction from a "gallery set" of images representing the environment. The poses of new "query" images can be estimated by matching them to the gallery set. The pose information can then be utilized for navigating to different points on the map. This process is illustrated in Fig.~\ref{fig:vloc_idea}.

Much of the ongoing research on visual localization focuses on developing methods that are more robust to viewpoint and appearance changes between the query and gallery images. In recent years, various benchmarking datasets have been published \cite{eth_zurich_computer_vision_group_and_microsoft_mixed_reality__ai_lab_zurich_eth-microsoft_2021,lee_large-scale_2021,sattler_benchmarking_2018}, and visual localization challenges have been hosted as part of the top-tier computer vision conferences. 
The new methods strive for even more accurate results on these benchmarks.

The most common applications of visual localization that are mentioned in the literature are autonomous driving and augmented reality \cite{lee_large-scale_2021,sattler_benchmarking_2018}. However, none of the new deep learning based localization methods have been demonstrated as part of a robot navigation stack. This raises the question: how relevant are the performance metrics used by the visual localization benchmarks and challenges for autonomous navigation? And how accurate do the localization methods actually have to be in order to enable autonomous robot navigation?

This paper seeks to address these concerns. We present a benchmark based on the Carla simulator~\cite{dosovitskiy_carla_2017} that enables testing visual localization methods for autonomous navigation. In the environment, the user can test how various state-of-the-art methods for visual localization perform when they are used for guiding the navigation of an autonomous car. This ability to directly test the performance of visual localization algorithms in their intended purpose enables the discovery of relevant new research problems, as compared to focusing on just measuring the algorithms' accuracy. Using a simulator also enables experiments that study the effects of factors such as illumination conditions, weather, and camera perspective on visual navigation.
Furthermore, the use of a simulator enables comparing the output of the visual localization algorithms with the accurate ground truth location of the autonomous vehicle, something that is not usually possible with real-world datasets~\cite{brachmann_limits_2021}. As pointed out by Brachmann \etal~\cite{brachmann_limits_2021}, synthetic data seems "easier" for a visual localization method to handle when compared to real camera data. Because of this, the navigation results reported in our work are likely to be overly optimistic. Even so, we argue that the ability to test an end-to-end visual navigation stack provides important new knowledge to the discussion on visual localization methods.

Our main contributions are: 
\textbf{1)} A simulator benchmark that enables development and evaluation of visual localization methods for autonomous navigation tasks;
\textbf{2)} An example use-case of the benchmark: evaluation of the state-of-the-art visual localization methods as part of a navigation stack;
\textbf{3)} Novel findings that connect an established visual localization performance metric, recall rate, with a proposed new metric, failure rate.
All results are fully reproducible and the benchmark is publicly available.

\section{Related Work}
\label{sec:related}
This work focuses on application of visual localization to {\em autonomous navigation}, which can be used by mobile robots to handle
various tasks such as delivery, inspection and people transportation~\cite{floreano_science_2015,utriainen_review_2018}. Vision-based navigation is useful in conditions where GPS or other sensors such as LiDAR~\cite{pfrunder_real-time_2017}, motion capture~\cite{michael_grasp_2010} or active localization beacons~\cite{macoir_uwb_2019} are not available or fail. The advantage of vision-based methods is that they only require commodity RGB-cameras that are cheap and power-efficient.

\paragraph{Vision-based navigation.}

There are various approaches to vision-based navigation. One of the most important factors differentiating the methods is the kind of prior information the navigating agent has about the environment it is operating in. The environment can be completely unknown, or the agent can have access to a a map representing the environment~\cite{anderson_evaluation_2018}.

In unknown environments a robot has to explore its surroundings. Its task can be to navigate to specific coordinates~\cite{chaplot_learning_2020}, find a certain object~\cite{chaplot_object_2020} or map the space~\cite{dang_graph-based_2020}.
For many applications, the ability to operate in known areas is sufficient~\cite{burgard_experiences_1999,furgale_stereo_2010}. In such cases, no exploration is needed. The robot can use cameras to determine its pose on a map representing the environment. The pose in turn can be used to plan a route to the robot's goal. The map can be a topological collection of images along the robot's route~\cite{dallosto_fast_2021}, a full metric map of the environment~\cite{oleynikova_real-time_2015}, or even implicitly encoded in an action policy derived by reinforcement learning~\cite{morad_embodied_2021}. In this taxonomy, visual localization falls under the group of methods that utilize metric maps. Visual localization has been utilized for navigation purposes in planetary rovers~\cite{furgale_stereo_2010}, wheeled utility robots~\cite{paton_bridging_2016} and drones~\cite{oleynikova_real-time_2015,warren_theres_2019}, for example.

\paragraph{Visual localization.}
%
There are various approaches to visual localization, such as pose regression~\cite{laskar_camera_2017}, scene coordinate regression~\cite{shotton_scene_2013} and direct image alignment~\cite{sarlin_back_2021,stumberg_gn-net_2020}, but in recent years {\em hierarchical localization}~\cite{irschara_structure--motion_2009,sarlin_coarse_2019} methods
have dominated the benchmarks.


Hierarchical localization consists of two stages. As a precondition, let's assume access to a "gallery" set of images representing the environment, from which a 3D reconstruction has been created by SLAM or SfM.
At first stage of localization, the gallery images most similar to a new "query" image are retrieved using place recognition methods~\cite{masone_survey_2021,zhang_visual_2021}. Then, local features extracted from the query image and the most similar gallery images are matched~\cite{ma_image_2021}. The real-world locations of the gallery features are known from the 3D reconstruction, so the resulting 2D to 3D correspondences enable estimating the 6-DoF pose of the query image using Perspective-n-Point (PnP) methods~\cite{gao_complete_2003}. The hierarchical visual localization approach has proven to be robust to changes in viewpoint and appearance, and is computationally feasible even for large-scale environments~\cite{sarlin_coarse_2019}. 


%
One of the characteristics of navigation which is relevant for visual localization is the sequential nature of the image data that robots' cameras capture. 
The continuous motion provides a strong prior that can be utilized in the prior retrieval stage by
retrieving best matching image descriptor sequences instead of individual images~\cite{milford_seqslam_2012,naseer_robust_2014}, by creating descriptors representing whole image sequences~\cite{garg_seqnet_2021} or by using the current pose estimate as a prior for topological localization~\cite{stenborg_using_2020}. At the local feature matching stage, the generalized camera model~\cite{pless_using_2003} enables estimating the camera trajectory from multiple images simultaneously~\cite{stenborg_using_2020}.
Kalman filters~\cite{brosh_accurate_2019}, particle filters~\cite{akai_simultaneous_2018} and graph-based methods ~\cite{oleynikova_real-time_2015,warren_theres_2019} can further process the pose estimates to enable sensor-fusion and outlier rejection.


\paragraph{Benchmarks.}
To the authors' best knowledge the work presented in this paper is the first dynamic benchmark where the components of
visual localization based navigation stack can be individually developed and evaluated in fair and reproducible manner.
Traditionally, performance of visual localization methods has been evaluated using static datasets of real images (~\ie Aachen Day-Night~\cite{sattler_benchmarking_2018},
Oxford RobotCar~\cite{maddern_1_2017},
CMU VL~\cite{badino_real-time_2012} and
Visual Localization Benchmark~\cite{sattler_benchmarking_2018}) and synthetic images (~\ie SimLocMatch~\cite{balntas_simlocmatch-benchmark_2021}, TartanAir~\cite{wang_tartanair_2020} and V4RL~\cite{maffra_real-time_2019} ). While these datasets enable evaluating the accuracy of visual localization, they do not provide insights into how well the methods are suited for navigation tasks.

Simulators, on the other hand, enable reproducible experiments with sufficiently realistic interactions. Several simulated benchmarks for vision-based navigation exist. The annual challenges~\cite{facebook_ai_research_habitat_nodate,stanford_vision_and_learning_lab_igibson_nodate} of the iGibson~\cite{xia_interactive_2020} and Habitat~\cite{savva_habitat_2019} simulators on tasks such as PointGoal and ObjectGoal navigation~\cite{anderson_evaluation_2018} are a good example. While they provide good platforms for evaluating agents' navigation performance, the benchmarks aren't tailored for the analysis of visual localization methods: the focus is on operation in unknown environments. Our proposed Carla-based benchmark is specifically aimed for evaluation and investigation of the performance of visual localization when applied to the context of autonomous navigation.






\section{Simulation Benchmark}
\label{sec:method}

Based on the discussion in Sec.~\ref{sec:related}, we identified a research gap in the application of visual localization for autonomous navigation. Visual localization is an active research topic in computer vision,
but methods are evaluated using static datasets and it is unclear how well the methods work when the visual localization output is used for closed-loop control. As a solution we present a benchmark which enables easy experimentation with different visual localization methods as part of a navigation stack.
The platform enables investigating various factors that affect visual localization and subsequent navigation performance, for example those listed in Table~\ref{tab:sim_capabilities}.

The benchmark is based on the Carla autonomous driving simulator~\cite{dosovitskiy_carla_2017} and our ROS2~\cite{maruyama_exploring_2016} port of the Hloc visual localization toolbox~\cite{sarlin_hloc_2021}. Carla was chosen because of its simplicity of use, relatively high level of photorealism and ROS2 support via the Carla ROS bridge module. ROS2 enables easy integration of the demonstrated visual localization package with different robotic platforms. We want to emphasize that Carla was chosen independent of its automotive application. The insights of this paper concern autonomous robot navigation in general, not just autonomous driving.

\begin{table}[b]
    \caption{Factors that affect visual localization performance and whether they are supported by our benchmark and demonstrated in this paper.}
  \centering

\resizebox{0.8\linewidth}{!}{

\begin{tabular}{llcc}
F\#
& Factor 
& Possible 
& Reported \\ 

\hline
F1 & Illumination               & \cmark   & \cmark \\
F2 & Weather                    & \cmark   & \cmark \\
F3 & Viewpoint changes          & \cmark   & \cmark \\
F4 & Scene structure            & \cmark   & \cmark \\
F5 & Time of year (seasons)     & \xmark   & \xmark \\
F6 & Camera placement (extr.)   & \cmark   & \xmark \\
F7 & Camera parameters (intr.)  & \cmark   & \xmark \\

F8 & Multiple cameras    & \cmark  & \xmark\\ %

F9 & Dynamic objects             & \cmark  & \xmark\\ 

F10 & Headlights             & \cmark  & \xmark\\ 

\hline
\end{tabular}
}


  \label{tab:sim_capabilities}
\end{table}

\subsection{ROS2 Visual localization package}
\label{sec:vloc_package}


In order to provide a generic visual localization interface to autonomous agents, we created the {\em ROS-Hloc Package}. It is a ROS2-wrapper for the Hloc
toolbox~\cite{sarlin_hloc_2021} that is a collection of state-of-the-art visual localization methods
and utility functions. 
The original toolbox is designed for
static image collections, but our ROS-Hloc extends it to
images arriving in a real-time stream.

The ROS-Hloc workflow is as follows. First, a gallery set is collected for each test environment. Inside the simulator this is achieved by driving a reference run with Carla's built-in autopilot. Along the route, images are captured by a camera attached to the vehicle. The images are taken at steady intervals, and saved to disk along with the exact camera pose.
After the gallery set has been captured, the images are processed to extract global and local feature descriptors. These are saved in a gallery database for queries.  
The 3D locations of the extracted local features are then estimated with the Colmap SfM library~\cite{schonberger_structure--motion_2016,schonberger_pixelwise_2016}.
Instead of running full SfM reconstruction we use point triangulation from known camera poses~\cite{hartley_multiple_2004}. This produces higher quality 3D scene models than reconstructions from unordered collections of images. In the simulator, acquiring the exact camera poses is trivial. In the real world, LiDAR SLAM methods can be employed in the mapping phase to ensure high-fidelity 3D models~\cite{caselitz_monocular_2016,lee_large-scale_2021}.


At inference time, the vehicle captures a query image which is sent to ROS-Hloc for pose estimation. 
First, the most similar gallery images are retrieved by place recognition. The retrieved gallery images are divided into spatial clusters using co-visibility clustering \cite{sarlin_coarse_2019}. Then, local feature matching is used for establishing query-to-gallery 2D-3D correspondences. The correspondences are used as inputs to the Perspective-n-Point (PnP) \cite{gao_complete_2003} solver provided by Colmap to produce 6DoF pose estimates for each cluster. The pose estimate from the cluster with the highest number of inlier 2D-3D correspondences is chosen as the final output of the visual localization pipeline. This pose estimate is forwarded to the agent's {\em motion planning and control stack} where it is used for producing steering commands.

Hloc includes various localization method options. There are two global descriptor methods, NetVLAD~\cite{arandjelovic_netvlad_2018} and Ap-GeM~\cite{revaud_learning_2019}, which we test in conjunction with the four supported local feature extractors, SIFT~\cite{lowe_distinctive_2004}, D2-net~\cite{dusmanu_d2-net_2019}, R2D2~\cite{revaud_r2d2_2019} and SuperPoint~\cite{detone_superpoint_2018}. With all methods besides SuperPoint we conduct the local feature matching by nearest neighbor search with ratio test~\cite{ma_image_2021} (NN-ratio). With SuperPoint, we utilize the SuperGlue matcher~\cite{sarlin_superglue_2020} instead.

\subsection{Vehicle motion planning \& control components}
\label{sec:planning_control}

In addition to ROS-Hloc for visual localization, the navigation stack needs two more components:
a {\em motion planner} and a {\em controller}. 
Motion planner is sub-divided into {\em global} and {\em local planners}.
Based on a route description, a global planner produces a set of waypoints from the vehicle's start position to its target. It is used in combination with a local planner that, at each timestep, finds the current closest waypoint and passes it as a subgoal to the controller. The controller consists of two proportional-integral-derivative (PID) controllers~\cite{ang_pid_2005}, one for
longitudinal and one for lateral control of the vehicle. Its purpose is to produce steering commands to move the vehicle towards the waypoint from local planner.
We use the global planner and controller from Carla, the local planner we implemented ourselves.

\vspace{-0.5\medskipamount}
\paragraph{Sensor fusion \& sequential processing.} 

Before being used for motion planning, the pose estimates from visual localization are first forwarded to a Kalman filter which fuses the estimates with measurements from a simulated wheel odometry sensor. The true values from the ideal odometry are injected with gaussian noise to make the sensor more realistic. The reason we fuse the visual pose estimates with wheel odometry data is that the PID-controller of the vehicle requires pose input at a high frequency, which is not achievable with the current state-of-the-art hierarchical visual localization systems. The wheel odometry also enables the vehicle to get estimates of its position when the environment is so degraded that the PnP solver of the visual localization pipeline cannot converge to a solution. The vehicle can navigate using pose information from just wheel odometry, but as the odometry measurements contain noise, the estimated pose accumulates error over time. This drift limits wheel odometry navigation only for short distances. Fusing the wheel odometry with visual localization effectively corrects the drift. We use the extended Kalman filter (EKF) implementation of the robot\_localization ROS package~\cite{moore_generalized_2016}. To make the localization system more robust to outliers, visual pose estimates with more than 20 meters deviation from the filter's current state are discarded. This scheme addresses the sequential nature of visual localization in autonomous navigation context at the pose level, and introduces a degree of temporal stability to the pose estimates. We do not add sequential processing to the prior and local feature matching stages. These would be interesting research topics, but we argue that single-image localization is an important starting point for investigating the applicability of visual localization for autonomous navigation.


%
\subsection{Evaluation scheme and performance metrics}\label{sec:metrics}

Performance measurement methodology is an important part of any benchmark. In this work, we wanted to bring together
visual localization and autonomous navigation, and therefore our
metrics should be meaningful to both fields.
Autonomous driving oriented visual localization performance metrics such as "probability distribution of certain distance driven without localization" \cite{porav_adversarial_2018} and "maximum open loop distance" \cite{clement_learning_2020} have been proposed, but whether they indicate success in autonomous navigation or have been invented to circumvent limitations of static datasets is unclear.
Navigation performance is often measured by the \emph{Success Rate (SR)}~\cite{wang_tartanair_2020} or \emph{Success weighed by Path Length (SPL)}~\cite{anderson_evaluation_2018}. In the context of our benchmark, SR would be measured by repeating $N$ episodes of the test route, and calculating the ratio between successful navigations of the route and the total number of episodes. SPL extends SR by additionally measuring the deviation from the shortest path to goal.
For the metrics used in our benchmark, we combine insights from the fields of visual localization~\cite{sattler_benchmarking_2018}, autonomous navigation~\cite{anderson_evaluation_2018} and visual object tracking~\cite{cehovin_visual_2016} (VOT). The performance evaluation needs of VOT are similar to ours, and there the evaluation methodologies have been investigated rigorously. Inspired by the works of Kristan \etal~\cite{kristan_novel_2016} and Cehovin \etal~\cite{cehovin_visual_2016}, we adopt two metrics that describe localization \emph{accuracy} and \emph{robustness}.


\vspace{-1\medskipamount}
\paragraph{Recall rate.}

The first aspect of performance is the accuracy of the visual localization method. This is an intuitive measure of how well a visual localization method performs under different conditions. For each experiment environment we conduct a test where a vehicle drives through the test route with visual localization running.
To produce comparable measurements of the accuracy of the visual localization methods, we don't use the estimates for navigation, but only measure their accuracy. For navigation the vehicle uses ground truth pose information from the simulator. Navigating based on the visual localization estimates, which can contain large errors, would lead to the vehicle driving a bit different route on each test run, affecting the repeatability of the accuracy measurements.
For each combination of experiment settings and localization methods we report the \emph{localization recall}, which was adopted from Sattler~\etal~\cite{sattler_benchmarking_2018}. We report the proportion of correct poses within three error thresholds: $<$ 0.25m, 2$^\circ$ (T1), $<$ 0.5m, 5$^\circ$ (T2) and $<$ 5m, 10$^\circ$ (T3). The distance of the estimated pose is compared to the pose of the vehicle at the moment the visual localization input image was taken. 

\vspace{-0.5\medskipamount}
\paragraph{Failure rate.}

The recall rate of a localization method does not fully describe its performance in autonomous navigation context. A high accuracy value can conceal infrequent but catastrophically large localization failures, which cause the vehicle to crash. The SR~\cite{wang_tartanair_2020} and SPL~\cite{anderson_evaluation_2018} are common metrics for this kind of experiments. However, they do not fit ours' well. In SR and SPL methodologies, an attempt to navigate the test route ends after the first navigation failure along the route: an otherwise easy test route with one difficult segment will result in the same SR or SPL as a test route that is difficult in all its parts.
As argued by Cehovin~\etal~\cite{cehovin_visual_2016}, it is more informative to measure success with re-initializations. If localization error causes the vehicle to crash, the vehicle and its EKF state are re-initialized back to the point along the route before the failure. This enables evaluating the localization performance along all segments of the test route. 

Similar to VOT~\cite{kristan_novel_2016,cehovin_visual_2016} we report the \emph{average failure rate} of navigation

\begin{equation}
    F = \frac{1}{N} \sum\limits_{i=0}^N {\frac{r_{i}}{L_{path}}}  \enspace ,
\end{equation}

where $N$ is the number of test episodes, $r_{i}$ is the number of re-initializations required to complete the route on test episode $i$ and $L_{path}$ is the length of the route in kilometers. 

To evaluate the failure rate, we conduct multiple episodes of the vehicle driving a predefined route while localizing based on sensor data. Since the autonomy stack of the vehicle is effectively using the visual localization to correct the wheel odometry drift \ie to improve navigation performance from that of a wheel odometry only based system, we also measure the performance of a vehicle navigating using wheel odometry only.
This defines a 'baseline' to which the navigation stack with visual localization should be compared. In easy conditions, the visual localization can be expected to improve navigation performance, while in degraded conditions the estimates can be very wrong and actually harm navigation performance.


\section{Experiments}
\label{sec:experiments}

To demonstrate the capabilities of the proposed simulator benchmark we performed experiments to compare the ability of different visual localization methods to cope with gallery-to-query shifts in illumination, camera viewpoint and weather. We want to emphasize that our benchmark is not limited to these factors, and enables further experiments with e.g. those listed in Table~\ref{tab:sim_capabilities}.

We limit our experiments to the state-of-the-art hierarchical localization methods, and do not consider other approaches such as direct image alignment ~\cite{stumberg_gn-net_2020,sarlin_back_2021} or sequence-based methods~\cite{milford_seqslam_2012,naseer_robust_2014}. Testing other method families would be an interesting research topic which we leave for future work.

\vspace{-1.0\medskipamount}
\paragraph{Environments.}

Out of the 8 default maps in Carla we selected two that represent
very different environments. Town01 is a model of a small town with densely packed buildings and a river in the center. We defined a ~1.2-kilometer long test route, from which a gallery set of 615 images was produced. The route consists of straight road segments connected by five 90 degree turns.
Town10 is a part of a bigger city with large buildings, skyscrapers and a beach. The route is approximately 0.5 kilometer long with six turns. The captured gallery set has 237 images.
Both the gallery sets were gathered by driving around the towns in mid-day sunny conditions and capturing images every 2 meters by a camera pointed perpendicular to the right of the vehicle's direction of travel. These gallery sets were used in all the subsequent experiments.

\begin{figure}[t]
  \centering
  \begin{adjustbox}{width=0.55\linewidth, valign=T}
   \includegraphics[width=1.0\linewidth, trim={5cm 0.cm 0cm 0cm},clip]{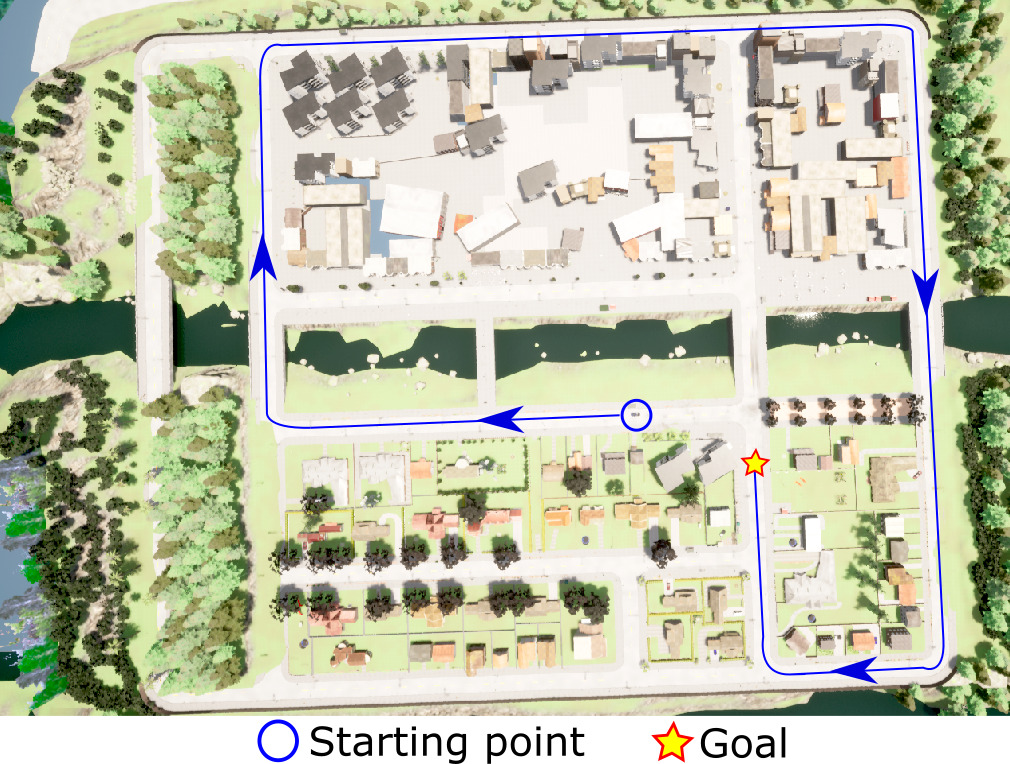}
   \end{adjustbox}
     \begin{adjustbox}{width=0.44\linewidth, valign=T}
   \includegraphics[width=1.0\linewidth, trim={4cm 2.cm 3.75cm 1.75cm},clip]{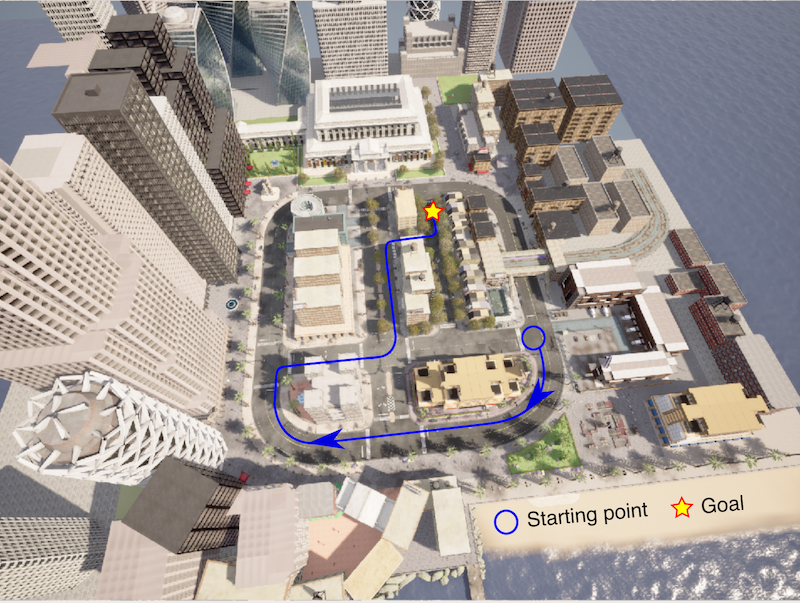} 
   \end{adjustbox}
   
   \vspace{-0.2cm}
   \caption{A bird's-eye view of the test routes in Town01 (left) and Town10 (right)}
      \label{fig:birdeye_towns}
      \vspace{-0.35cm}
\end{figure}


\vspace{-2.0\medskipamount}
\paragraph{Methods and common parameters.}
Each of the localization methods included in Hloc (see Sec.~\ref{sec:vloc_package}) were tested in all the experiments. None of the methods were retrained with data from the simulator; pretrained model weights were acquired from the original Github repositories. The number of gallery images retrieved by place recognition for pose estimation was set to 5. Ratio threshold of NN-ratio matcher was set to 0.8, and SuperGlue was used with the default parameters. For all of the localization methods we used input image resolution of 800x600 pixels. The target speed of the vehicle and the localization frequency were set to $4 m/s$ and $2 Hz$, respectively. Magnitude of the wheel odometry noise was set to a level which causes the pose estimate to drift away from the true pose at a rate of $8.5\%$ for the position and $0.4^\circ/m$ for the orientation.


\subsection{Illumination change}
\label{sec:illumination_change}

This experiment evaluated the performance of the state-of-the-art visual localization methods under query-to-gallery illumination change in "Town01" and "Town10" (Fig.~\ref{fig:birdeye_towns}).
The autonomous vehicle was set to perform the test routes under multiple illumination conditions (Fig.~\ref{fig:short}). In the easiest test scenario the illumination corresponds to that of the gallery set, and in the most difficult scenarios there is almost complete darkness. We report the average failure rate for 5 repetitions of the test route for each localization method and illumination condition.
For the recall rate, we only report the results from one run per test condition since the vehicle drives using an error-free controller and therefore the variance between the runs is negligible.

In Carla, illumination is controlled by two parameters: sun intensity and sun elevation angle. These two have to be adjusted jointly to produce a full range of illumination conditions from daylight to darkness. We start from sun intensity value of $1.0$ and elevation angle $0.4\pi$ $rad$, and for each successive illumination condition we halve the values from the previous condition by parameter $k$:
\begin{equation}
    V_k = V_{base} * 0.5^k
    \label{eq:illumination}
\end{equation}
where $V_{base}$ is sun intensity or elevation angle value for the gallery set, and $V_k$ is the elevation angle or intensity for illumination condition $k$. $k \in [0,1,...,10]$ leads to 11 distinct illumination conditions (see Fig.~\ref{fig:short} for examples).

\subsection{Viewpoint change}
\label{sec:viewpoint_change}

In this experiment, a viewpoint difference was introduced between the gallery and query images. The query images view the same scene content as the gallery, but from a different perspective. This was implemented by introducing a series of vertical offsets $z$ and pitch angle decreases $\theta$ to the pose of the localization camera in the test runs (Fig.~\ref{fig:short}).
We report average failure and recall rates of the different offsets, tested in Town01 under illumination level $k=0$. Other settings were kept the same as in the illumination change experiments in Sec.~\ref{sec:illumination_change}.

\subsection{Weather change}
\label{sec:weather_change}

\begin{figure}[b]
  \centering
  
  \begin{subfigure}[t]{0.32\linewidth}
    \includegraphics[width=0.98\linewidth]{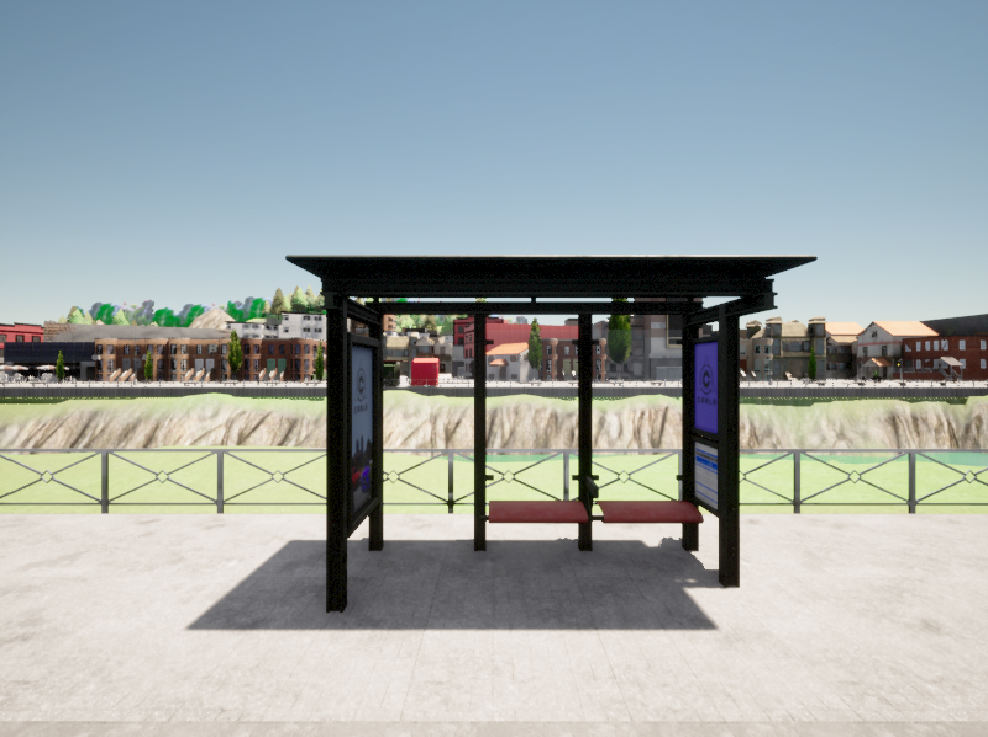}
    \caption{$k=0$}
    \label{fig:short-a}
  \end{subfigure}
  \begin{subfigure}[t]{0.32\linewidth}
    \includegraphics[width=0.98\linewidth]{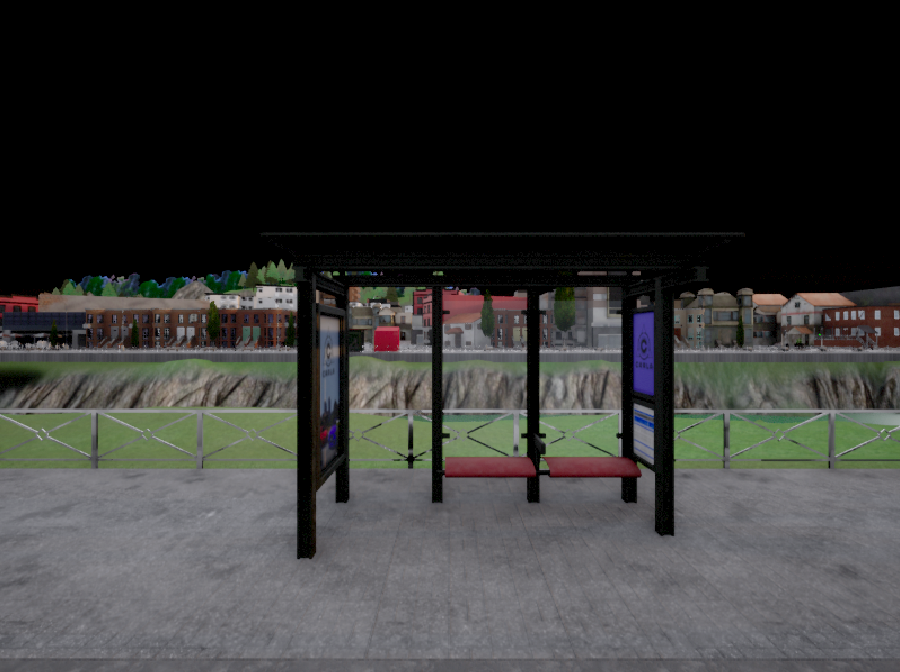}
    \caption{$3$}
    \label{fig:short-b}
  \end{subfigure}
  \begin{subfigure}[t]{0.32\linewidth}
    \includegraphics[width=0.98\linewidth]{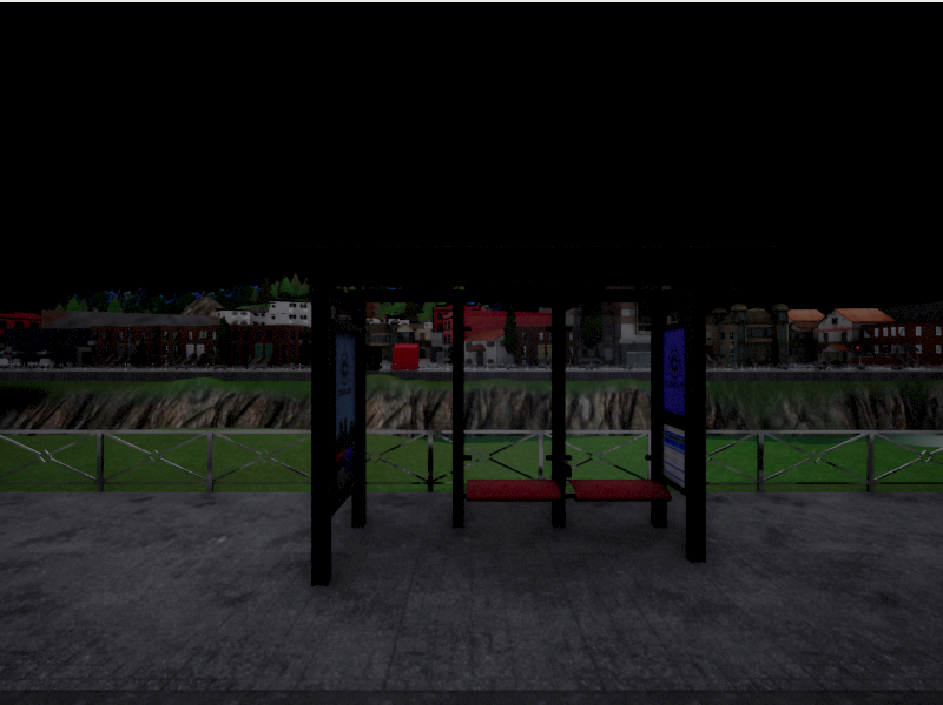}
    \caption{$6$}
    \label{fig:short-c}
  \end{subfigure}
  
    \begin{subfigure}[t]{0.32\linewidth}
    \includegraphics[width=0.98\linewidth]{source/figures/town01_start_k0.png}
    \caption{$z= 0~m,\, \theta = 0^\circ$}
    \label{fig:short-d}
  \end{subfigure}
    \begin{subfigure}[t]{0.32\linewidth}
    \includegraphics[width=0.98\linewidth]{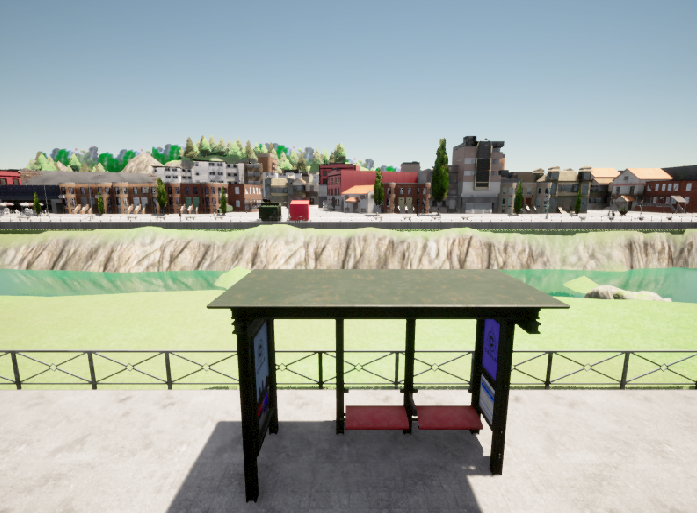}
    \caption{$2~m,\, 10^\circ$}
    \label{fig:short-e}
  \end{subfigure}
  \begin{subfigure}[t]{0.32\linewidth}
    \includegraphics[width=0.98\linewidth]{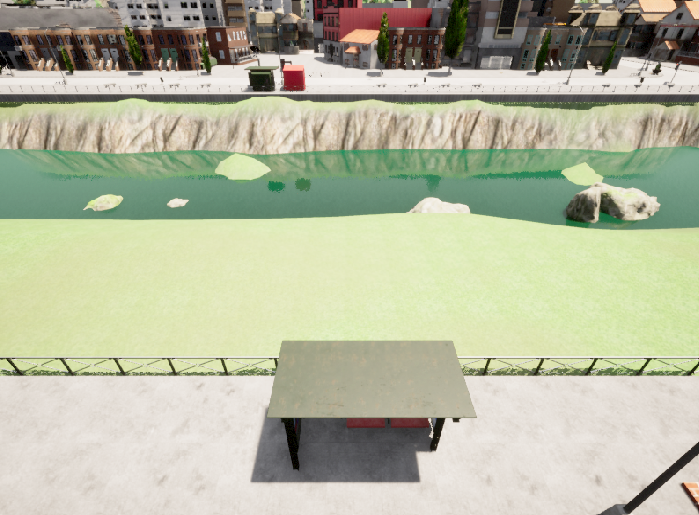}
    \caption{$7~m,\, 35^\circ$}
    \label{fig:short-f}
  \end{subfigure}

    \begin{subfigure}[t]{0.32\linewidth}
    \includegraphics[width=0.98\linewidth]{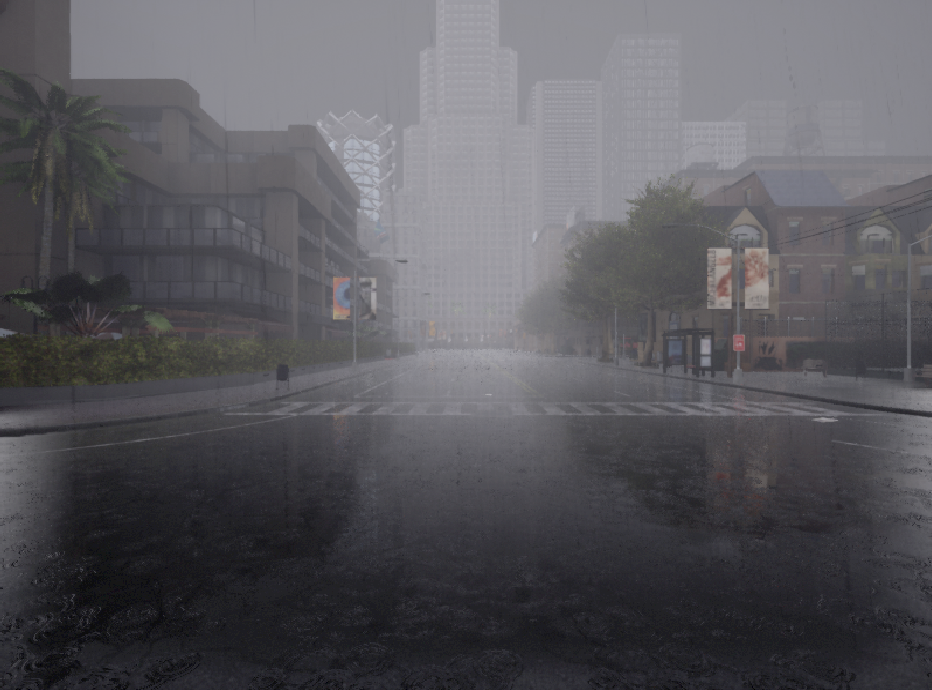}
    \caption{$v=90~m$}
    \label{fig:short-g}
  \end{subfigure}
    \begin{subfigure}[t]{0.32\linewidth}
    \includegraphics[width=0.98\linewidth]{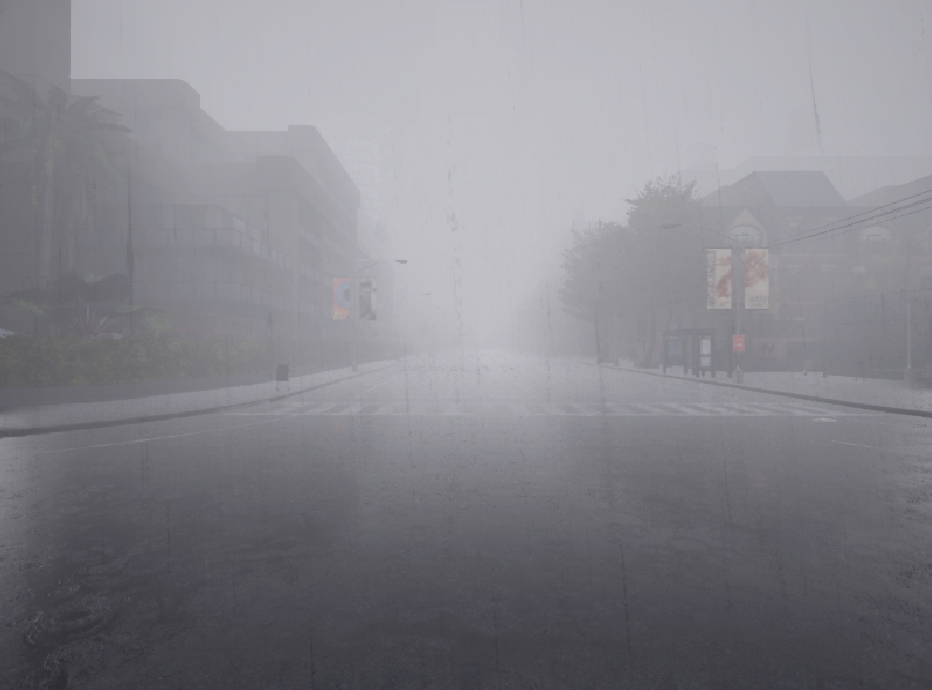}
    \caption{$60~m$}
    \label{fig:short-h}
  \end{subfigure}
  \begin{subfigure}[t]{0.32\linewidth}
    \includegraphics[width=0.98\linewidth]{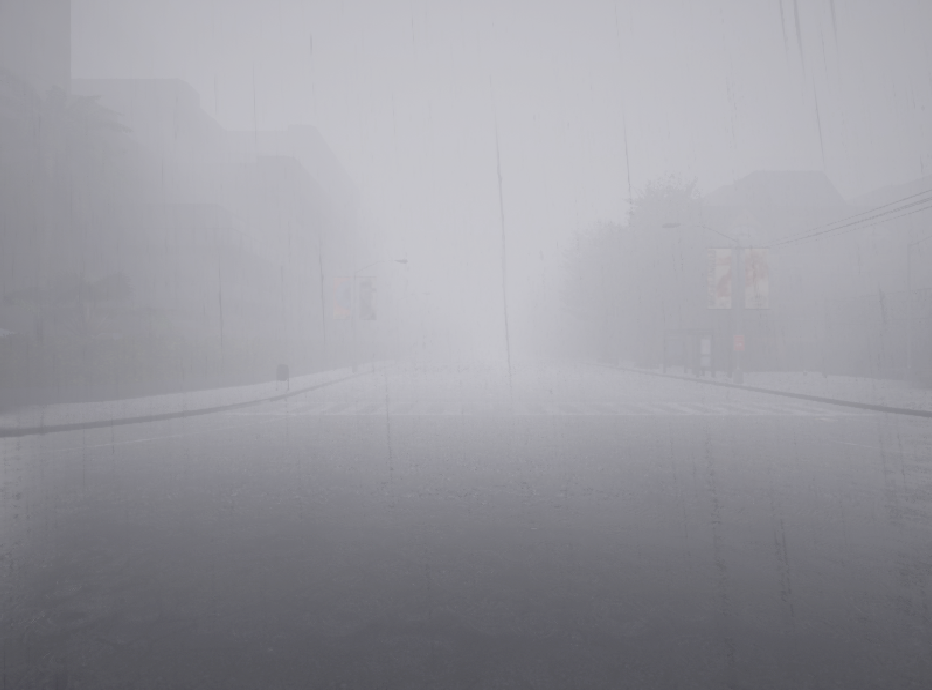}
    \caption{$10~m$}
    \label{fig:short-i}
  \end{subfigure}
  \vspace{-0.1cm}
  \caption{Town01 under different illumination (a-c) and viewpoint (d-f) shifts, and weather changes in Town10 (g-i)}
  \label{fig:short}
\end{figure}

We conducted additional experiments on the effect of gallery-to-query weather change in Town10. Illumination level was set to $k=0$ and rain was added into the environment. Then, we introduced progressively increasing amounts of fog. The amount is controlled by defining how close to the vehicle the fog begins. We created 4 weather conditions with visual ranges $v \in [90.0, 60.0, 30.0, 10.0]$ meters (Fig.~\ref{fig:short}). We report the failure and recall rates over 5 repetitions of each condition and method.

\section{Results and Discussion}
\label{sec:results}


\begin{figure*}[ht]
  \centering
     \begin{subfigure}[t]{0.33\linewidth} 
   \includegraphics[width=1.0\linewidth, trim={0 1.1cm 0cm 1cm}]{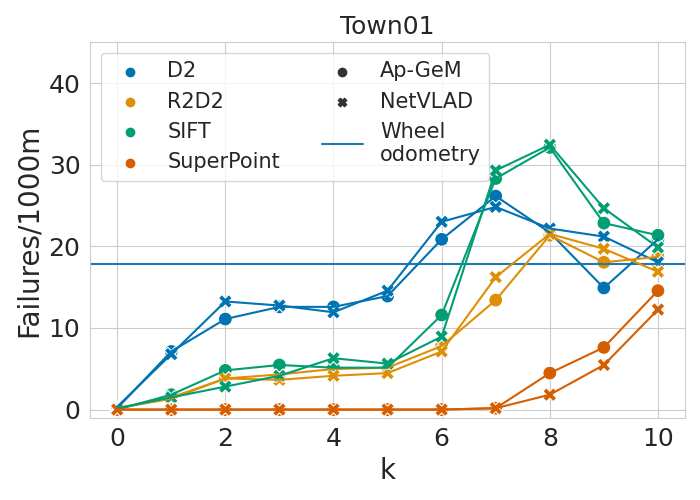}
   \caption{}
    \label{fig:failure_ill_town1}
   \end{subfigure}
     \begin{subfigure}[t]{0.33\linewidth} 
   \includegraphics[width=1.0\linewidth, trim={0 1.1cm 0cm 1cm}]{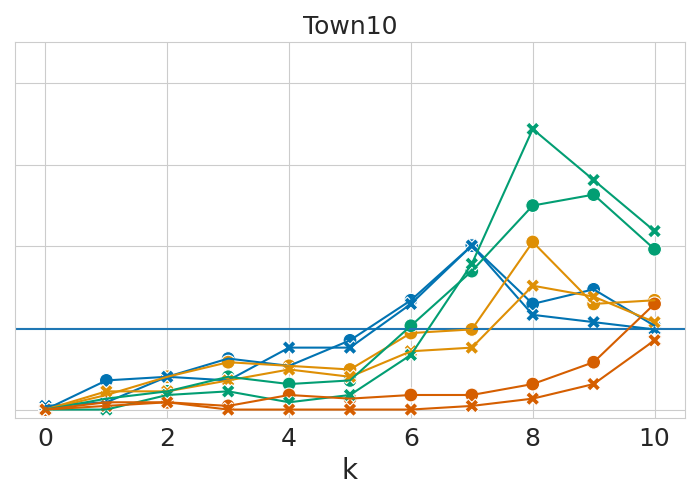}
      \caption{}
    \label{fig:failure_ill_town10}
   \end{subfigure}
  \begin{subfigure}[t]{0.33\linewidth} 
     \includegraphics[width=1.0\linewidth, trim={0 1.1cm 0cm 1cm}]{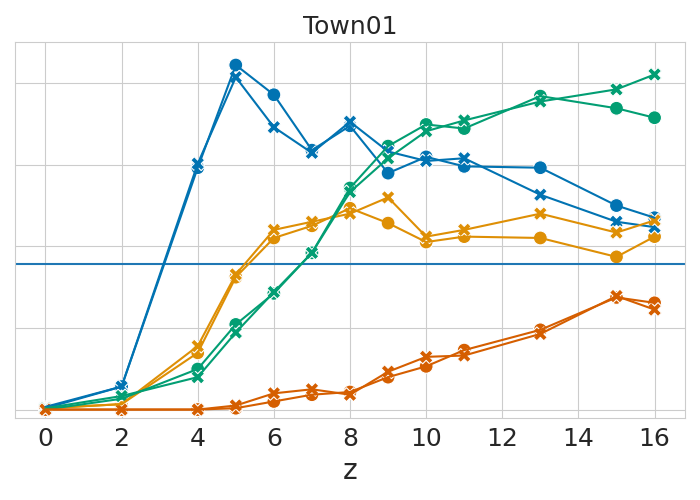}
    \caption{}
    \label{fig:failure_pitch_town1}
  \end{subfigure}
  \vspace{-0.25cm}
   \caption{ Relationship of failure rate with illumination (\ref{fig:failure_ill_town1},~\ref{fig:failure_ill_town10}) and viewpoint change (\ref{fig:failure_pitch_town1}). Marker color indicates type for local features, shape for global features.}
   \label{fig:fr_kvalue_correlation}
   \vspace{-0.1cm}
\end{figure*}

\begin{table*}[!t]
\caption{Navigation failure rates over 5 repeated runs of the same route in each daylight illumination level $k$ conditions. Smaller is better. PR $=$ place recognition method, LF $=$ local feature type, CT $=$ computation time (\textit{ms}).}
\vspace{-0.25cm}
  \centering
  \resizebox{0.90\linewidth}{!}{
  \begin{tabular}{l l r r r r r r r r r r r r  |  r r r r r r r r r r r | c }
  &  & \multicolumn{11}{c}{Town01} &  \multicolumn{13}{c}{Town10}       \\
    \cmidrule(lr){3-25}
     PR & LF &  $k=$ 0 & 1 & 2 & 3 & 4 & 5 & 6 & 7 & 8 & 9 & 10 & &
     0 & 1 & 2 & 3 & 4 & 5 & 6 & 7 & 8 & 9 & 10 & CT
     \\
    \midrule
    Ap- & Sift  & \textbf{0.0} & 1.8 & 4.8 & 5.5 & 5.1 & 5.1 & 11.6 & 28.3 & 32.1 & 22.8 & 21.4 & 
    &     \textbf{0.0} & 1.3 & 2.2 & 4.0 & 3.1 & 3.6 & 10.3 & 17.0 & 25.0 & 26.3 & 19.6
    & 169 
    \\
     GeM & D2-net  & 0.2 & 7.1 & 11.1 & 12.6 & 12.6 & 13.9 & 20.9 & 26.2 & 21.7 & 14.9 & 20.9 & 
    &  \textbf{0.0} & 3.6 & 4.0 & 6.2 & 5.4 & 8.5 & 13.4 & 20.1 & 12.9 & 14.7 & 10.3
    & 165 
     \\
     & R2D2  & 0.2 & 1.5 & 3.8 & 4.3 & 5.0 & 5.1 & 7.8 & 13.4 & 21.4 & 18.0 & 18.7 & 
    & \textbf{0.0} & 1.8 & 4.0 & 5.8 & 5.4 & 4.9 & 9.4 & 9.8 & 20.5 & 12.9 & 13.4
    & 194 
     \\
     & SuperPoint & \textbf{0.0} & \textbf{0.0} & \textbf{0.0} & \textbf{0.0} & \textbf{0.0} & \textbf{0.0} & \textbf{0.0} & \textbf{0.2} & 4.5 & 7.6 & 14.6 &
     &  \textbf{0.0} & 0.9 & \textbf{0.9} & 0.4 & 1.8 & 1.3 & 1.8 & 1.8 & 3.1 & 5.8 & 12.9
    & 193 
     \\
     
    \midrule
    Net- & Sift  & 0.2 & 1.5 & 2.8 & 4.1 & 6.3 & 5.6 & 8.9 & 29.3 & 32.5 & 24.7 & 19.9 &
    &  \textbf{0.0} & \textbf{0.0} & 1.8 & 2.2 & 0.9 & 1.8 & 6.7 & 17.9 & 34.4 & 28.1 & 21.9
    & \textbf{134}  
    \\
     VLAD & D2-net  & 0.3 & 6.8 & 13.2 & 12.7 & 11.9 & 14.6 & 23.0 & 24.8 & 22.2 & 21.2 & 18.0 &
    &  0.4 & 0.9 & 4.0 & 3.6 & 7.6 & 7.6 & 12.9 & 20.1 & 11.6 & 10.7 & 9.8
    & 139 
    \\
     & R2D2  & 0.2 & 1.3 & 3.8 & 3.6 & 4.1 & 4.5 & 7.1 & 16.2 & 21.5 & 19.7 & 16.9 &  
     &  \textbf{0.0} & 2.2 & 2.2 & 3.6 & 4.9 & 4.0 & 7.1 & 7.6 & 15.2 & 13.8 & 10.7
    & 167 
     \\
     & SuperPoint  & \textbf{0.0} & \textbf{0.0} & \textbf{0.0} & \textbf{0.0} & \textbf{0.0} & \textbf{0.0} & \textbf{0.0} & \textbf{0.2} & \textbf{1.8} & \textbf{5.5} & \textbf{12.3} & 
     &  \textbf{0.0} & 0.4 & \textbf{0.9} & \textbf{0.0} & \textbf{0.0} & \textbf{0.0} & \textbf{0.0} & \textbf{0.4} & \textbf{1.3} & \textbf{3.1} & \textbf{8.5}
    & 166  
     \\
     \midrule
     \multicolumn{2}{l}{Wheel odometry} &  \multicolumn{11}{c}{17.9}  & & \multicolumn{10}{c}{9.8} & &
      \\
     \bottomrule
  \end{tabular}
  }
  \label{tab:failurerates}
  \vspace{-0.2cm}
\end{table*}

\begin{figure}[b]
  \centering
  \begin{adjustbox}{width=0.95\linewidth, valign=T} 
     \includegraphics[width=1.0\linewidth, trim={0 1cm 0cm 1cm}]{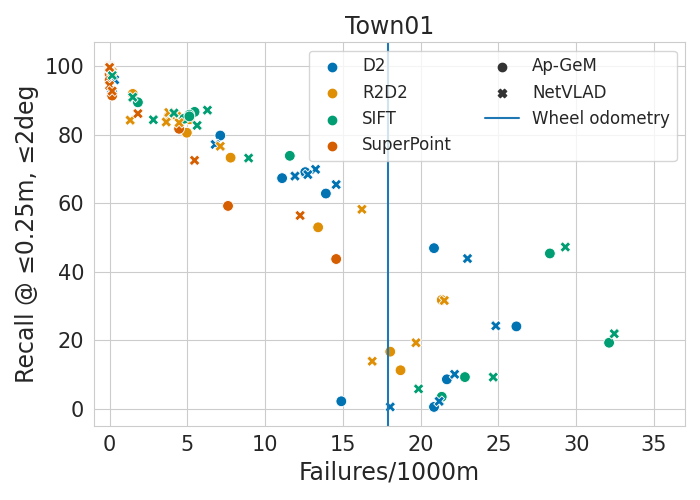}
  \end{adjustbox}
   \caption{Relationship between the failure rate and recall rate T1. Marker color and shape indicate feature type. } 
   \label{fig:fr_recall_correlation}
\end{figure}




\begin{figure*}[ht]
 
%
%
    \begin{minipage}[t]{0.69\linewidth}
    \begin{minipage}[t]{\linewidth}
    \captionof{table}{Navigation failure rates over 5 repetitions of the same route at each gallery-to-query camera pose (viewpoint) offset. $z=$ elevation shift, $\theta=$ pitch shift.}
    \vspace{-0.2cm}
    \centering 
    \resizebox{0.92\linewidth}{!}{ 
  \begin{tabular}{l l r r r r r r r r r r r r r r }
  & & \multicolumn{13}{c}{Town01} &  \\
    \cmidrule{3-15}
     \multirow{2}{*}{PR} & \multirow{2}{*}{LF} &  $z=$ 0 & 2 & 4 & 5 & 6 & 7 & 8 & 9 & 10 & 11 & 13 & 15 & 16 &

     \\

      &  &  $\theta=$ 0 & 10 & 22.5 & 27.5 & 32.5 & 35 & 37.5 & 40 & 40 & 40 & 40 & 40 & 40 &

     \\
    \midrule
    Ap- & Sift & \textbf{0.0} & 1.3 & 5.0 & 10.4 & 14.2 & 19.2 & 27.2 & 32.3 & 34.9 & 34.4 & 38.4 & 36.9 & 35.8 & 

    \\
     GeM & D2-net  & 0.2 & 2.8 & 29.6 & 42.2 & 38.6 & 31.8 & 34.8 & 29.0 & 31.0 & 29.8 & 29.6 & 25.0 & 23.5 & 
     \\
     & R2D2  & 0.2 & 0.7 & 7.0 & 16.2 & 21.0 & 22.5 & 24.7 & 22.8 & 20.5 & 21.2 & 21.0 & 18.7 & 21.2 & 
     \\
     & SuperPoint & \textbf{0.0} & \textbf{0.0} & \textbf{0.0} & \textbf{0.2} & \textbf{1.0} & \textbf{1.8} & 2.2 & \textbf{4.0} & \textbf{5.3} & 7.3 & 9.8 & \textbf{13.7} & 13.1 & 
     \\
     
    \midrule
    Net- & Sift  & 0.2 & 1.7 & 4.0 & 9.4 & 14.4 & 19.2 & 26.7 & 30.8 & 34.1 & 35.4 & 37.7 & 39.2 & 41.1 &
    \\
     VLAD & D2-net  & 0.3 & 2.8 & 30.1 & 40.7 & 34.6 & 31.5 & 35.3 & 31.6 & 30.5 & 30.8 & 26.3 & 23.0 & 22.4 &
    \\
     & R2D2  & 0.2 & 0.7 & 7.8 & 16.6 & 22.0 & 23.0 & 24.0 & 26.0 & 21.2 & 22.0 & 24.0 & 21.7 & 23.2 &
     \\
     & SuperPoint  & \textbf{0.0} & \textbf{0.0} & \textbf{0.0} & 0.5 & 2.0 & 2.5 & \textbf{1.8} & 4.6 & 6.5 & \textbf{6.6} & \textbf{9.3} & 13.9 & \textbf{12.3} &
     \\
     \midrule
     \multicolumn{2}{l}{Wheel odometry} &  \multicolumn{12}{c}{17.9} 
      \\
     \bottomrule
  \end{tabular} 
  \label{tab:pitch_failures}
  }
  \end{minipage}%
  \hfill%

\vspace{1.0em}
%
%
\begin{minipage}[t]{\linewidth}
\captionof{table}{The localization recall rates for the reference paths with thresholds T1 ($\le$ 0.25m, $\le$2$^\circ$), T2 ($\le$0.50m,$\le$5$^\circ$) and T3 ($\le$5.00m, $\le$10$^\circ$). Table with all values of $k$ in the appendix.}\label{tab:accuracies} 
\vspace{-0.2cm}
  \centering
  \resizebox{1.0\linewidth}{!}{
  \begin{tabular}{l l r r r r r r r}
    & \multirow{2}{*}{PR} & \multirow{2}{*}{LF} & $k=$ 0 & 2 & 4 & 6 & 8 & 10  \\
    &  &   & T1 / T2 / T3 & T1 / T2 / T3 & T1 / T2 / T3 & T1 / T2 / T3 & T1 / T2 / T3 & T1 / T2 / T3 \\
    \cmidrule(lr){2-9}
   \parbox[t]{2mm}{\multirow{8}{*}{\rotatebox[origin=r]{90}{Town01}}}
       
   &  Ap- & Sift   & 98.0 / 98.2 / 99.8 & 84.7 / 89.6 / 96.5 & 85.9 / 89.3 / 96.7 & 73.8 / 78.5 / 90.5 & 19.2 / 23.8 / 29.8 & 3.5 / 5.8 / 8.6     \\
    
    & GeM & D2-net   & 92.3 / 95.7 / 99.8 & 67.3 / 74.7 / 90.1 & 69.1 / 74.7 / 88.0 & 46.9 / 58.1 / 73.8 & 8.6 / 10.9 / 15.6 & 0.5 / 0.8 / 2.0   \\
     
   &  & R2D2   & 98.0 / 98.4 / \textbf{100.0} & 85.5 / 90.4 / 97.7 & 80.6 / 88.2 / 97.0 & 73.3 / 79.7 / 93.6 & 31.7 / 37.5 / 50.5 & 11.2 / 12.9 / 15.0    \\
     
   &  & SuperPoint  & \textbf{100.0} / \textbf{100.0} / \textbf{100.0} & \textbf{99.8} / \textbf{99.8} / 99.8 & 99.5 / \textbf{100.0} / \textbf{100.0} & \textbf{95.9} / 98.5 / 99.0 & 81.7 / 86.3 / 90.5 & 43.7 / 48.9 / 57.9   \\
    \cmidrule(lr){2-9}

  & Net- & Sift   & 97.4 / 98.5 / 99.8 & 84.4 / 87.7 / 97.5 & 87.2 / 89.8 / 97.4 & 73.2 / 79.6 / 92.3 & 21.9 / 27.0 / 34.4 & 5.8 / 8.2 / 13.0      \\
    
    & VLAD & D2-net   & 96.1 / 96.9 / 99.7 & 69.9 / 75.8 / 90.1 & 67.9 / 75.7 / 87.3 & 43.8 / 52.4 / 75.0 & 10.0 / 13.0 / 17.9 & 0.5 / 1.2 / 1.8      \\
     
    & & R2D2   & 98.4 / 98.5 / 99.7 & 86.5 / 91.0 / 97.5 & 86.2 / 89.5 / 97.5 & 76.6 / 82.4 / 93.1 & 31.6 / 37.7 / 49.2 & 13.8 / 15.3 / 18.3      \\
     
    & & SuperPoint   & \textbf{100.0} / \textbf{100.0} / \textbf{100.0} & 99.7 / 99.7 / \textbf{100.0} & \textbf{99.8} / \textbf{100.0} / \textbf{100.0} & 94.4 / \textbf{99.2} / \textbf{99.5} & \textbf{86.2} / \textbf{90.8} / \textbf{96.1} & \textbf{56.4} / \textbf{60.4} / \textbf{68.1}  \\
     
     \cmidrule(lr){2-9} \\
    \cmidrule(lr){2-9}
    \parbox[t]{2mm}{\multirow{8}{*}{\rotatebox[origin=r]{90}{Town10}}}
       & Ap- & Sift  &   
         95.2 / 96.4 / 99.6 & 89.5 / 90.3 / 92.7 & 87.1 / 87.9 / 91.5 & 76.6 / 79.0 / 83.9 & 12.5 / 16.5 / 33.5 & 0.0 / 0.4 / 3.6
        \\
    
   & GeM & D2-net   &  
    94.7 / 98.4 / 99.2 & 84.2 / 88.7 / 91.9 & 81.9 / 87.1 / 92.7 & 46.8 / 53.6 / 65.3 & 0.0 / 0.0 / 1.2 & 0.0 / 0.0 / 0.0
     \\
     
    & & R2D2   & 
     93.1 / 94.7 / 99.6 & 88.3 / 90.7 / 92.3 & 85.5 / 87.9 / 90.3 & 75.8 / 78.2 / 83.5 & 15.7 / 21.4 / 31.5 & 0.4 / 0.8 / 2.8
     \\
     
   &  & SuperPoint   &   
    99.6 / \textbf{100.0} / \textbf{100.0} & 96.0 / 96.0 / 96.0 & 94.4 / 94.4 / 94.4 & 93.1 / 93.5 / 93.5 & 73.4 / 75.4 / 76.2 & 35.9 / 37.5 / 41.5
     \\
    \cmidrule(lr){2-9}
    
   & Net- & Sift   &   
    96.4 / 97.2 / \textbf{100.0} & 92.7 / 94.0 / 94.8 & 87.9 / 90.7 / 93.5 & 75.4 / 77.8 / 85.1 & 12.9 / 17.3 / 35.5 & 0.4 / 0.4 / 2.0
    \\
    
    & VLAD & D2-net   &   
      97.2 / 99.6 / \textbf{100.0} & 86.7 / 89.5 / 94.0 & 83.1 / 88.3 / 92.7 & 45.2 / 58.5 / 69.4 & 0.0 / 0.0 / 0.8 & 0.0 / 0.0 / 0.0
     \\
     
   &  & R2D2  &   
     91.6 / 92.8 / 99.6 & 88.7 / 91.1 / 92.7 & 85.9 / 89.1 / 90.7 & 76.6 / 80.6 / 87.1 & 14.1 / 20.2 / 33.9 & 0.4 / 0.8 / 2.0
     \\
     
   &  & SuperPoint   &  
       \textbf{100.0} / \textbf{100.0} / \textbf{100.0} & \textbf{97.2} / \textbf{97.2} / \textbf{97.6} & \textbf{97.6} / \textbf{98.0} / \textbf{98.0} & \textbf{98.0} / \textbf{98.0} / \textbf{98.0} & \textbf{86.7} / \textbf{87.9} / \textbf{88.7} & \textbf{50.0} / \textbf{55.2} / \textbf{58.1}
     \\
     
     \cmidrule(lr){2-9}
          \cmidrule(lr){2-9}
  \end{tabular}
}
\end{minipage}%
\end{minipage}%
\hfill%
%
  %
  \begin{minipage}[t]{0.30\linewidth}
  \begin{minipage}[t]{\linewidth}
  \captionof{table}{Failure rates at gallery-to-query weather (visibility) changes $v$.}%
  \vspace{-0.2cm}
    \centering
    \resizebox{0.9\linewidth}{!}{ 
  \begin{tabular}{l l r r r r }
   & &\multicolumn{4}{c}{Town10}  \\
    \cmidrule{3-6}
    PR & LF & $v=$ 90 & 60 & 30 & 10 

     \\
    \midrule
    Ap- & Sift &  8.5 & 25.4 & 25.4 & 21.9  

    \\
    GeM & D2-net  & 12.1 & 13.8 & 10.3 & 12.9 
     \\
     & R2D2  & 10.3 & 12.5 & 10.3 & 10.3 
     \\
     & SuperPoint & \textbf{0.0} & 0.4 & 1.8 & 2.2 
     \\
     
    \midrule
    Net- & Sift  & 7.6 & 25.9 & 29.5 & 25.9 
    \\
     VLAD & D2-net  & 12.5 & 13.4 & 10.7 & 9.4
    \\
     & R2D2  & 9.4 & 8.0 & 7.6 & 8.5  
     \\
     & SuperPoint  & \textbf{0.0} & \textbf{0.0} & \textbf{0.0} & \textbf{0.9} 
     \\
     \midrule
     \multicolumn{2}{l}{Wheel odometry}  & \multicolumn{4}{c}{9.8} 
      \\
     \bottomrule
  \end{tabular}
  }
  \label{tab:weather_failures}
  \end{minipage}%
  \hfill%
\vspace{0.75em}
%
%
\hfill%
\begin{minipage}[t]{\linewidth} 
\centering
 \begin{adjustbox}{width=0.75\linewidth, valign=t}
 \hfill%
   \includegraphics[width=1.0\linewidth, trim={4cm 0.75cm 3.25cm 0.5cm},clip]{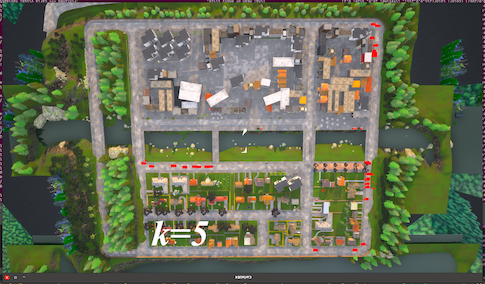}
   \hfill%
   \end{adjustbox}
   \hfill%

     \begin{adjustbox}{width=0.75\linewidth, valign=t}
     \hfill%
   \includegraphics[width=1.0\linewidth, trim={4cm 0.75cm 3.25cm 0.5cm},clip]{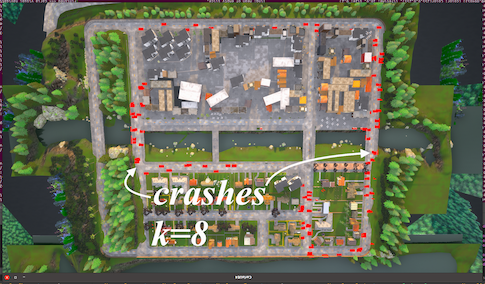}
   \hfill
   \end{adjustbox}
   \vspace{-0.2cm}
   \captionof{figure}{Crash locations (red) for NetVLAD + R2D2 in Town01.} \label{fig:crash_sites}
\end{minipage}%
\end{minipage}
\vspace{-0.9cm}
\end{figure*}


Here we provide a thorough analysis of the illumination experiment results, and for compact presentation show only the most important findings for the two other experiments: viewpoint and weather change. Full result tables and additional visualizations are provided in the appendix, available in the supplementary material.

\subsection{Illumination change}

\paragraph{Failure rates.}
Table~\ref{tab:failurerates} shows the navigation failure rates for the illumination experiments. The same data is visualized in Fig.~\ref{fig:fr_kvalue_correlation}.
As expected, the failure rates stay low when the gallery-to-query illumination shift is small. The rates rise with increasing severity of the shift. After exceeding the odometry failure rate, the rate for each method peaks and then starts to decrease. Around the peak, the gallery-to-query appearance change is large enough to cause big errors in the visual localization. However, they are not as large as to cause rejection by the EKF $20m$ outlier threshold. As the shift further grows, more pose estimates are rejected by the filter, and the vehicle starts mainly relying on wheel odometry. As result, the failure rate decreases and converges towards that of wheel odometry only.

Combinations using SuperPoint achieve the lowest failure rates, and by a clear margin. The ability of SuperPoint to improve navigation performance is remarkable. The method brings benefits over wheel odometry even at $k=9$, when localizing the images is very difficult even for the human eye.
The best performing combination is that of SuperPoint and NetVLAD, followed by SuperPoint with Ap-GeM. This follows a general pattern: The local feature method seems to have more effect on the performance than the place recognition method. Of the two place recognition methods, NetVLAD provides a slightly better performance. R2D2 and SIFT are consistently tied for the second best local feature. D2, a deep learning based feature, ranks the worst. The good performance of SIFT is interesting: published in 1999, it can still compete with the new methods.

\vspace{-1\medskipamount}
\paragraph{Town comparison.} 
The failure rates exhibit some differences between the two towns. In Town01, SuperPoint has significantly better performance than the other methods. In Town10, this gap is more narrow. This is likely caused by a higher degree of perceptual aliasing in the scene, such as buildings with repetitive textures. However, in both environments the overall order of method performance remains approximately the same - only R2D2 and SIFT switch places on some values of $k$.

\vspace{-1\medskipamount}
\paragraph{Visual localization recall vs.~failure rate.} 

Table~\ref{tab:accuracies} presents the recall rates for the illumination experiments. The extent to which recall, a visual localization performance metric, measures navigation performance is an important question. Fig.~\ref{fig:fr_recall_correlation} shows the correlation between the two metrics for the illumination experiments. As expected, the correlation is strong, but the plot also provides two important findings:
\textbf{1)} SuperPoint, that achieves the lowest failure rates in different illumination conditions, also achieves the lowest failure rate for a given recall rate;
\textbf{2)} there is a certain operation point, determined by odometry drift, after which changes in the recall rate become meaningless for autonomous navigation. Especially the second finding is interesting as it shows that visual localization performance needs to be sufficiently good in order to improve over wheel odometry only.
For Town01, the recall rate of a method at threshold T1 has to be above 60\% to benefit navigation.
In other words, improving recall from 40\% to 50\% is almost meaningless while improvement from 60\% to 70\% is clearly significant. This is intuitive, but not possible to observe from static datasets.

\vspace{-1.5\medskipamount}
\paragraph{Spatial distribution of navigation failures.}
The failure rate metric describes how often navigation failures happen, but it doesn't describe \emph{where} the failures happen. The simulator enables visualizing the failure locations and identifying 
segments that are difficult for visual localization. Fig.~\ref{fig:crash_sites} shows one example of such visualization.


\vspace{-1\medskipamount}
\paragraph{Method runtimes.} 

Table~\ref{tab:failurerates} also shows the average runtimes of each method, measured from Town10. These delays induce noise into the motion planning: a pose estimate is used for control after a small lag, during which the vehicle has moved a small distance from the point for which the pose estimate has been computed. At a target velocity of $4m/s$, the distances are in the range of $[0.54, 0.78]$ meters. This noise, however, is mostly in the longitudinal direction and doesn't seem to drastically affect navigation performance. During turns, the effect is more dominant.

\subsection{Viewpoint and weather change}


\paragraph{Viewpoint change}
The results for the viewpoint change are in Table~\ref{tab:pitch_failures}, and visualized in Fig.~\ref{fig:failure_pitch_town1}.
The ranking of the methods is the same as in the previous illumination change experiment. Interestingly, the performance gap between SuperPoint and the other methods is even greater in the viewpoint experiments. At $z=7$, none of the other methods help navigation, but even at $z=16$ SuperPoint performance still hasn't deteriorated to level of wheel odometry. In the illumination experiments, this gap wasn't as wide. SuperPoint seems especially robust to viewpoint change, which is important for navigation applications.

\vspace{-1.5\medskipamount}
\paragraph{Weather change.}
The failure rates of the weather experiments in Table~\ref{tab:weather_failures} do not follow the same trend as the illumination and viewpoint experiments. 
Only SuperPoint shows correlation between the visual range and failure rate - it is also the only method to improve navigation performance at any value of $v$. Even the easiest condition $v=90m$ is very difficult for the other methods. Investigating what creates robustness to this kind of gallery-to-query variation could be an interesting topic for future research.

\subsection{Summary}
The extensive experiments over illumination, viewpoint and weather change, and in two different maps,    show that the best navigation performance is achieved by SuperPoint paired with either of the two place recognition methods, NetVLAD or Ap-GeM. SuperPoint performs well even in the tough conditions presented in Fig.~\ref{fig:short} \textit{(c), (f)} and \textit{(i)} while the other methods do not improve navigation over wheel odometry at such severe gallery-to-query changes.

The in-depth analysis of the illumination experiment verifies 
the utility of the proposed new metric, failure rate.
We believe that the proposed format of analysis proves useful to the development of navigation-oriented visual localization methods.

\section{Conclusion}
\label{sec:conclusion}

This paper introduced a simulator benchmark for testing and developing visual localization methods as a part of a vision-based autonomous navigation stack. To demonstrate the capabilities of the benchmark we evaluated popular visual localization methods 
under gallery-to-query appearance and viewpoint changes.
The results show that the benchmark and the proposed navigation failure rate metric can reveal information about the visual localization methods that is not evident from the traditional static benchmarks. 
Substantial differences were observed in the performances of the methods, and it is evident that some are better suited to vision-based navigation than others.

In the future, the benchmark could be used for studying the performance gap between single-image and sequential visual localization, or for investigating the effect of factors such as camera placement. We hope that the research community finds the proposed benchmark useful for finding new, exciting research directions for vision-based autonomous navigation.

\paragraph{Acknowledgements.} This research has received funding from the Technology Innovation Institute (TII) as part of the ARROWSMITH project.




{\small
\bibliographystyle{ieee_fullname}
\bibliography{references.bib}
}


\end{document}